\documentclass[runningheads]{llncs}
\usepackage{amsmath}
\usepackage{amssymb}
\usepackage{algpseudocode}
\usepackage{algorithm}
\usepackage{enumitem}
\usepackage{multirow} 
\usepackage{booktabs} 
\usepackage{xcolor}
\usepackage{soul}
\usepackage[T1]{fontenc}
\usepackage{graphicx}
\begin{document}
\title{Local-to-Global Logical Explanations for Deep Vision Models}
%
%
\author{Bhavan Vasu\inst{1}\orcidID{0000-0003-4961-580X} \and
Giuseppe Raffa\inst{2}\orcidID{0009-0003-9178-4559} \and
Prasad Tadepalli\inst{1}\orcidID{0000-0003-2736-3912} 
}
\authorrunning{B. Vasu et al.}
%
%
\institute{Oregon State University OR 97331, USA \\
\email{\{vasub, tadepall\}@oregonstate.edu} \and
Intel Labs, Hillsboro. Oregon, USA \\
\email{giuseppe.raffa@intel.com}}
\maketitle              
\newcommand{\remove}[1]{}

\begin{abstract}
While deep neural networks are extremely effective at classifying images, they remain opaque and hard to interpret. 
We introduce local and global explanation methods for black-box  models that generate explanations in terms of human-recognizable primitive concepts.  
Both the local explanations for a single image and the global explanations for a set of images 
are cast as logical formulas in monotone disjunctive-normal-form (MDNF), whose satisfaction guarantees that the model yields a high score on a given class. We also present an algorithm for explaining the classification of examples into multiple classes in the form of a  
monotone explanation list over primitive concepts. Despite their simplicity and interpretability we show that the explanations maintain high fidelity and coverage with respect to the blackbox models they seek to explain in challenging vision datasets. 

\keywords{Explainable AI  \and Neurosymbolic AI \and Monotone DNF \and Deep Learning}
\end{abstract}

\section{Introduction}
\label{sec:intro}

Deep visual models now equal or surpass human accuracy on benchmark
classification tasks. Yet, their internal logic remains opaque, which raises questions of transparency and accountability of AI decision making.  
The UK and EU General Data Protection Regulation (GDPR) grants
data subjects the right to be informed about “the logic involved’’ in
automated decision making in critical applications. 
Our two-level interpretability framework maps naturally onto the strata
of stakeholders envisaged by the GDPR. The \emph{local} explanation for a single image provides an instance-specific rationale in object space—precisely capturing 
the “concise and intelligible’’ narrative required when a person
asks, “why was \textit{my} photo labeled this way?’’. While auditors and internal compliance teams desire a class-level  \emph{global} summary that
      supports statistical auditing: coverage tells what fraction of the
      data is governed by the global explanation, enabling checks for
      systematic bias or hidden spurious cues.
By delivering explanations at two complementary granularities, the
proposed approach addresses the GDPR’s right-to-explanation spectrum from individual transparency to organizational accountability, without
sacrificing the proprietary status of the underlying model.

Designing fully interpretable models often requires costly feature engineering in domains like computer vision and still delivers only modest accuracy. We instead construct logical, post‐hoc explanations for high‐performing deep models, thereby combining performance with high level interpretability. Our approach produces both local and global explanations by leveraging annotated samples from the model’s target distribution. We partition these samples into a support set for deriving global, rule‐based explanations and a validation set—for objectively evaluating explanation quality and coverage. To satisfy the legal requirements of GDPR and enable widespread acceptance, a practical explanation system for a given model must satisfy two distinct desiderata:

\begin{enumerate}[label=(\roman*),nosep,leftmargin=*]
\item \textbf{Local insight}  
      Given a \emph{fixed} pair \((x,y)\), where $x$ is an image and $y$ is a class, the system should say
      \emph{why} the model assigns $y$ to $x$. The explanation is \emph{not} a prediction, but a reason for the prediction in that it identifies which parts of $x$ are responsible for the prediction. 

\item \textbf{Global summary}  
      Domain experts need a global picture: Which recurring patterns of the image data does the model rely on and how much of the dataset does each pattern cover?
\end{enumerate}

We base our paper on two key observations. First, the confidence score of a model's prediction on a given image typically grows monotonically with positive evidence of that class in the image. Second, there are often multiple pieces of evidence, each of which is strongly and independently predictive of the class \cite{shitole2021one}. The above observations lead us to a perturbation-based search framework that allows us to compute multiple local explanations. But unlike in the work of \cite{shitole2021one}, where the explanations are subimages, 
each explanation here is expressed as a logical conjunction of a minimal set of
positive literals that correspond to human-annotated 
objects or object parts in the image
whose presence is sufficiently 
Following \cite{darwiche2022computation}, we call
the set of all such minimally sufficient explanations, which can be expressed as a disjunction of conjunctions of positive literals, a {\em complete explanation} of the deep model's classification of that image.  Although our method assumes object information, it can be obtained automatically: in non-generalizable domains, annotating a small subset of images and applying label-propagation methods \cite{vasu2025beyond} suffices, while in generalizable domains, foundational segmentation models such as Segment Anything \cite{carion2025sam} can generate object masks without manual supervision.

A complete explanation of a model's classification of an image in terms of object parts can thus be viewed as a logical formula in monotone disjunctive normal form (MDNF) \cite{abasi2014exact} that captures the model's behavior on
that image. The monotone property means that all primitive literals are positive and correspond to the relevant object parts in the image. Intuitively, the model's classification is sufficiently explained by the objects present in {\em any one} of the conjunctions in the formula. On the other hand, the absence of at least one object from each of the
conjunctions would reduce the
confidence score of the model below the  desired threshold for the predicted class. 
Moving beyond local explanations, we seek a global explanation that succinctly summarizes the reasons that multiple images in a dataset are classified into a certain class. We express these reasons as an unordered list, called a {\em covering global explanation}, wherein 
each clause corresponds to a local explanation 
shared by multiple images and expressed in terms of object parts.

Unfortunately the covering explanations of a class might be too general and not sufficiently discriminating in that a single explanation in terms of object parts might cover images of different classes. To avoid this ambiguity and simultaneously explain multiple classes, we introduce a {\em global multiclass explanation} in the form of an explanation list, where each clause of the list is a conjunction of positive literals that maps to a class. The explanation list uniquely maps each image into a class supported by one of its local explanations and can be viewed as a summary of explanations for all classes over the entire dataset.

We make the following contributions. 
\begin{itemize}[nosep,leftmargin=*]
\item A local explanation algorithm based on beam search that computes  the complete concept-level local 
explanation for images and their predicted classes. 
\item A global explanation algorithm based on greedy set cover that succinctly explains the classifications of all examples of a given class. 

\item A multi-class global explanation algorithm that simultaneously explains the classification of a set of examples into multiple classes. 
 
\item Experimental evaluation of our methods in multiple vision datasets and multiple deep models.
\end{itemize}
In the following sections, we will formally introduce our approach and provide results that support the above contributions.

\section{Related Work}
Most prior interpretation methods for vision models focus on pixel‐level saliency: e.g.\ LIME \cite{zhang2019should}, RISE \cite{DBLP:conf/bmvc/PetsiukDS18}, or classical saliency‐map techniques \cite{zhou2016learning,selvaraju2020grad,fong2017interpretable}, which lack semantic grounding and suffer from user‐dependent bias. Consequently, these local masks cannot be reliably combined into global explanations.  Some concept‐level methods \cite{bau2017network} require white‐box access to model weights, whereas we treat the classifier as an opaque scoring oracle.  Moreover, recent studies \cite{shitole2021one} demonstrate that a single image may admit multiple valid rationales, exacerbating aggregation challenges.  To address this, we adopt a concept‐based approach that aligns local explanations with a human taxonomy, providing a stable, canonical basis for both local and global interpretation. After extraction, the local concepts become literals in a multi–clause Boolean formula in disjunctive normal form (DNF).  
While tools like SHAP \cite{mosca2022shap} or LIME could, in principle, treat each primitive as a feature, LIME relies heavily on negative evidence and models the explanation as a surrogate classifier. Inspired from global explainability methods on tabular data \cite{darwiche2022computation} and graphs 
\cite{azzolin2023global}, we propose a novel global explanation method for image classification.
\par Early rule-extraction work such as \textsc{TREPAN} converts a trained network into a decision tree by recursively querying the model, but the tree grows in feature space and offers no guarantee that a leaf’s antecedent is \emph{sufficient} for the prediction \cite{craven1995extracting}. Modern neuro-symbolic pipelines go a step further by \emph{building}
logical structure into the network: DL2 \cite{fischer2019dl2} imposes first-order constraints during training, LogicMP \cite{xulogicmp} and related layers embed Horn clauses directly in the forward pass, and
Concept-Bottleneck Models \cite{koh2020concept} learn an explicit concept lattice that must be annotated or invented a-priori to training the deep model. More recently DNF-MT\cite{baugh2025neural} even co-optimises a neural policy and its DNF explanation end-to-end. While these systems produce symbolic artifacts aiming to explaining a deep model, they either (i) require
concept supervision before training, (ii) modify the optimization objective, or (iii) restrict the hypothesis class to tractable logic during training. Our method is complementary: it treats the network as a \emph{black-box oracle}, searches post-hoc for minimal sufficient concept level explanations, and compiles them into a global  explanation list only after model training. Consequently we (a) need no concept
labels before training, (b) leave the original objective or accuracy of the model untouched, and (c) evaluate faithfulness through the sufficiency of explanations. The result is a symbolic explanation whose clauses are guaranteed sufficient on the instance level and jointly exhaustive on the class level, providing a model-agnostic alternative to architecture-dependent neuro-symbolic designs.

\section{Sufficient and Complete Local  Explanations}
\begin{figure}
    \centering
    \includegraphics[width=\linewidth]{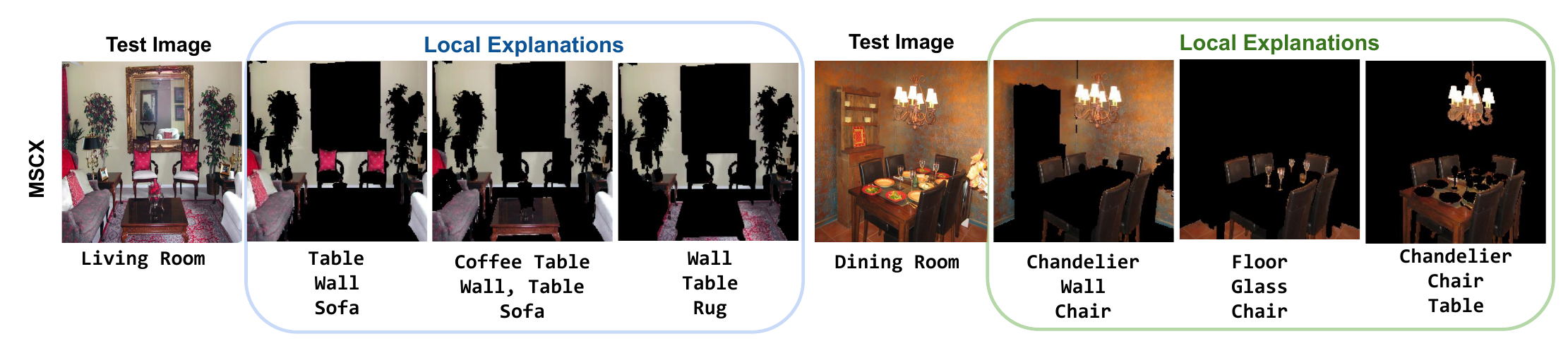}
    \caption{Figure shows the visualization of multiple minimally sufficient concept based local explanations (MSCX) $P^{min}(x, y)$ for two images from the classes {\em Living Room} and {\em Dining Room} from the ADE20k dataset. Different local explanations focus on different sets of objects.}
    \label{fig:mscx}
\end{figure}

Let $X \subseteq \mathbb{R}^{H \times W \times 3}$ and 
  $Y = \{1,\dots,C\}$
denote the spaces of RGB images and class labels, respectively. Let $f$ be a neural network model we would like to explain. We formalize $f$ 
as a mapping
\[
  f : X \longrightarrow \mathbb{R}^{C},
  \qquad
  x \longmapsto \bigl(f_{1}(x),\dots,f_{C}(x)\bigr),
\]
and write \(f_{y}(x)\) for the scoring function associated with class \(y\in Y\). When employed as a classifier $f$ maps image $x$ to the class label 

$y^* = {\rm arg}\max_y f_y(x)$. 

\emph{Object–wise decomposition:} Instead of explaining the classifier's workings simply with a saliency map of the input image, we seek to build explanations in terms of symbolic primitives. In this work, we assume that every image \(x\in X\) consists of a set of non‑overlapping image segments that correspond to human-interpretable objects or parts, denoted by
$O(x) = \{o_{1},\dots,o_{k}\}$. \\
\textbf{Masking and Blurring:} To build an explanation of $f$ on any input $x$, we allow the explainer an oracle access to $f$ on $x$ and its perturbations obtained by masking and blurring parts of the image. For any object subset \(S\subseteq O(x)\) let 
\(x_{S}\) be the
\emph{masked} image obtained by blurring all pixels outside the
objects in \(S\), that is, keeping only pixels in \(S\). Since masking solely based on the raw pixels in $S$ can in turn give away the object being masked, 
we run mask $S$ through a gaussian filter before masking, to create a smoother transition between pixels in $S$ and pixels not in $S$.\\
\textbf{Explanation Space:} In the past it has been observed that 
the presence of a small set of relevant patches in the image is often sufficient for the neural network models to give a high score for a given class \cite{shitole2021one}. To make these visual explanations more symbolic and transferable across images, 
in this paper, we consider
explanations in terms of the afore-mentioned human-interpretable objects as shown in \ref{fig:mscx}. 
In particular, 
for a sufficiency ratio \(\tau_{P}=0.95\), we define the set of sufficient explanations of an image \(x\) and class \(y\) to be the sets of objects defined by: 
\begin{align}
  P(x,y)
    &:=\Bigl\{\,S\subseteq O(x)\setminus\{\varnothing\}\;
              \Bigl|\;
              f_{y}\bigl(x_{S}\bigr)\;\ge\;\tau_{P}\,f_{y}(x)\Bigr\}
  \label{eq:important}
\end{align}

Equation~\eqref{eq:important} designates as sufficient any combination of objects 
whose pixels alone preserve at least $\tau_{P}$ of the original
\(f_{y}(x)\). 
\emph{Minimally Sufficient and Complete Explanations:} Building on the insight of \cite{shitole2021one}, we note that the score $f_y(x_S)$ typically grows monotonically with $S$, as more object parts of the image are exposed. This entails that if a set $S$ is in $P(x,y)$, then all its supersets are also in. 
Hence, $P(x,y)$ contains many redundant explanations, which can be eliminated for succinctness. 

We say a set \(S\in P(x,y)\) is \emph{minimally sufficient} if no strict
subset of $S$ is in  $P(x, y) $:

\[
   P_{\min}(x,y)
     \;:=\;
     \Bigl\{\,S\in P(x,y)\;\Bigl|\;
            \nexists\;T\subsetneq S,
            T\in P(x,y)\Bigr\}
\tag{MI}
\]

Thus, each member of \(P_{\min}(x,y)\) represents a set of object or parts which are minimally sufficient to classify the instance $x$ into class $y$ by the model.  Following \cite{darwiche2022computation}, 
who called such construct a ``complete reason,'' we call 
\(P_{\min}(x,y)\) a {\em complete explanation} for classifying an example $x$ into class $y$. Intuitively, all other reasons or explanations are  subsumed by those in this set.

\remove{PT: I believe the rest of the discussion can be eliminated.
}

\remove{

\[
   m^{\star}_{x,y}\;:=\;|P_{\min}(x,y)|
   \qquad\text{and index}\qquad
   P_{\min}(x,y)
     =\bigl\{\tilde{S}^{(i)}_{x,y}\bigr\}_{i=1}^{m^{\star}_{x,y}}
\]

\paragraph{Non‑important sets.}
All remaining non‑empty subsets are labelled \emph{non‑important}:
\[
    R(x,y)\;:=\;
      \bigl(2^{O(x)}\setminus\{\varnothing\}\bigr)\setminus P(x,y)
\]
Consequently
\[
    P(x,y)\;\cup\;R(x,y)=2^{O(x)}\setminus\{\varnothing\},
    \qquad
    P(x,y)\cap R(x,y)=\varnothing 
\]

Assume the image $x$ is decomposed into $k:=|O(x)|$ atomic patches.
The set of \emph{all non‑empty combinations} is the power set
\[
    \mathcal{P}(x)\;=\;2^{O(x)}\setminus\{\varnothing\},
\qquad
    |\mathcal{P}(x)|\;=\;2^{k}-1 
\tag{C1}
\]
}

\remove{
\subsection{Problems with existing methods}
\emph{TODO: Local and Global approaches clean up}
\paragraph{Time complexity.}
Any algorithm that evaluates \emph{independently} on each subset $S\subseteq O(x)$ requires at least one forward pass of
$f(\cdot)$ per subset.  Therefore the \emph{runtime lower bound} is
\[
    T(k)\;=\;\Omega\!\bigl(2^{k}\bigr),
\tag{C2}
\]
which grows exponentially.  For a modest $k=32$ objects
this already means ${\approx}\,4.3\times10^{9}$
forward evaluations which is utterly impractical. 

\paragraph{Memory complexity.}
If the algorithm stores the decision/logit for every subset,
the \emph{space} requirement is
\[
    M(k)\;=\;\Theta\!\bigl(2^{k}\bigr)\quad\text{bits},
\tag{C3}
\]
since each of the $2^{k}-1$ subsets must be stored as
“important’’ or “non‑important’’.
For $k=32$ this is gigabytes even with a single
bit per subset; for higher‐resolution
partitionings it becomes astronomically large. Additionally, finding the \emph{minimal} sufficient sets
$\tilde{S}\in P_{\min}(x,y)$ can be cast as a
\emph{minimal hitting‑set} or \emph{set cover} variant,
which is NP‑hard \cite{karp2009reducibility}.
Thus, even clever pruning cannot avoid exponential worst‑case behaviour unless $\mathrm{P}=\mathrm{NP}$.

\par Because exhaustive search is infeasible, we propose the use of a heuristic search with beam search, to approximate important subsets. Hence the proposed framework defines importance via Equation (1) but in practice relies on approximate algorithms to discover an approximate representative subset of $P_{\min}(x,y)$. }

\section{Computing Complete Explanations}

We can represent a complete explanation as a propositional monotone DNF formula.
Each member of the complete explanation \(P_{\min}(x,y)\)  can be viewed as a conjunction of positive literals, each of which represents the presence of an object part. Since any one of these sets is sufficient to classify $x$ into $y$, the set of all such sets can be viewed as a disjunction of conjunctions that entails class $y$. Since the formula contains only positive literals, it falls under the special class of monotone DNF 
formulae. Finding \(P_{\min}(x,y)\) that fully explains the classification of $x$ by model $f$ can thus be viewed as learning a monotone DNF formula by asking membership queries on perturbations of input $x$. This problem 
has been well-studied with the best algorithms almost matching their lower bounds \cite{abasi2014exact}. However, the sample complexities of the best algorithms and the lower bounds are still exponential in either the maximum size of the conjunctions in the target formula or the number of disjunctions. Hence we adopt a beam search algorithm,
which takes advantage of the continuous score output by the neural network model to guide the search and outputs a heuristic approximation of \(P_{\min}(x,y)\). 

Beam search is a space-efficient heuristic search algorithm that often yields near-optimal results. In the current context, it can be formulated
as a search starting with full image and incrementally removing pixels or starting with an empty image and incrementally adding patches that correspond to annotated objects. Since our goal is to approximate $P_{min}(x, y)$ with minimally sufficient sets of objects, we use the latter strategy which keeps the search space small. 

Let \(x_{\varnothing}\) denote the image obtained from \(x\) by blurring
(or zeroing) \emph{all} objects in \(O(x)=\{o_{1},\dots,o_{k}\}\).
Fix \(\tau_{P}=0.95\) and beam width \(B\).

\begin{algorithm}[H]
\caption{\textsc{BeamAdd}\((x,y,B,\tau_{P})\)}
\label{alg:beam-add}
\begin{algorithmic}[1]
\Require image \(x\), class \(y\), beam width \(B\)
\Ensure family \(P_{\min}(x,y)\) of $\tau_{P}$-minimal sufficient masks
\vspace{2pt}
\State $f^* \gets f_{y}(x)$ \Comment{reference logit}
\State $F \gets \bigl\{\varnothing\bigr\}$           \Comment{frontier (subsets already added)}
\State $\mathcal{S}_{\text{suf}}\gets\varnothing$    \Comment{collected sufficient sets}
\While{$F\neq\varnothing$}
    \State $C\gets\varnothing$                       \Comment{next-depth candidates}
    \ForAll{$S\in F$}
        \ForAll{$o\in O(x)\setminus S$}              \Comment{add one new object}
            \State $T\gets S\cup\{o\}$
            \If{$f_{y}\!\bigl(x_{T}\bigr)\ge\tau_{P}f^{*}$}
                \State $\mathcal{S}_{\text{suf}}\gets\mathcal{S}_{\text{suf}}\cup\{T\}$
            \Else
                \State $C\gets C\cup\{T\}$
            \EndIf
        \EndFor
    \EndFor
   
    \State $F\gets\text{top-}B\text{ subsets in }C\text{ by }f_{y}(x_{T})$ \Comment{beam pruning}
\EndWhile
\State \Return 
   \(
     P_{\min}(x,y)=
       \bigl\{\,T\in\mathcal{S}_{\text{suf}}
          \;\bigl|\;
          \nexists\,T'\subsetneq T:
              f_{y}(x_{T'})\ge\tau_{P}f^{*}
       \bigr\}
   \)
\end{algorithmic}
\end{algorithm}
Algorithm \ref{alg:beam-add} can be divided into four key stages enumerated below:
\begin{enumerate}[label=\textbf{\arabic*.},leftmargin=*]
\item \textbf{Initialisation (lines 1–3).}  
      The algorithm starts from the \emph{fully blurred} image
      \(x_{\varnothing}\) by placing the empty object set
      \(\varnothing\) in the frontier \(F\) (line 2).

\item \textbf{Expansion (lines 5–13).}  
      At each depth every subset \(S\) in the current frontier is
      expanded by \emph{adding exactly one new object}
      \(o\in O(x)\setminus S\) (lines 7–8).  
      The resulting set \(T=S\cup\{o\}\) is evaluated once through the
      model oracle.  
      If \(T\) already satisfies the $\tau_{P}$ sufficiency test
      (line 9) it is moved to the collector
      \(\mathcal S_{\text{suf}}\); otherwise it enters the candidate pool
      \(C\) for the next depth (line 11).

\item \textbf{Beam pruning (line 16).}  
      If no candidate at the current depth passes the threshold, the
      algorithm keeps only the \(B\) subsets in \(C\) that achieve the
      highest score, discarding the rest.  
     
\item \textbf{Redundancy removal (Line 18).}  

      A final filtering pass (return block) deletes any set that has a
      proper subset still in \(\mathcal S_{\text{suf}}\). 
      This step is needed since objects may not have been added in optimal order by the beam search, and some objects added in the previous steps might be redundant after other objects are added.      
      
\end{enumerate}

\textbf{Complexity:} At depth \(d\) each frontier set has \(d\) elements and \(|o(x)|-d\) possible
extensions. The size of the search frontier \(|F|\le B\). If 
$d^{\ast}$ is the maximum number of objects in any minimally sufficient concept based local explanations (MSCX) and $k$ is the maximum number of objects in any image,
the time complexity is 
\(\mathcal O\!\bigl(B\,d^{\ast}\,k\bigr)\), which is
linear in beam width, maximum size of any MSCX, and object count.

\begin{theorem}
Every member of the list returned by Algorithm 1 satisfies the sufficiency test and is non-redundant. 
Further, if $f_y(x_S)$ is monotonic and the beam width is sufficiently large, then the algorithm correctly computes \( P_{min}(x,y)\). 
\end{theorem}

\noindent {\em Proof:}
Note that every set passes the sufficiency test in step 9 of the algorithm before it is inserted into $S_{suf}$. All redundant 
sets in $S_{suf}$ are pruned in step 18. 
This guarantees sufficiency and non-redundancy of the output. Further if the beam width is sufficiently large, the algorithm reduces to breadth first search, which preserves all subsets in the search space. Although supersets are not generated for the sets in $S_{suf}$, they are redundant when $f_y(x_S)$ is monotonic. 
$\Box$

\remove{
Let \(d^{\ast}\) be the \emph{minimum} 
number of objects that must be added to 
reach the $\tau_{P}$ threshold. Because the search begins with the empty set and
explores subsets in non-decreasing size, 
the first depth at which a candidate passes the threshold is exactly \(d^{\ast}\).  
Every set collected in line 6 therefore has cardinality \(d^{\ast}\); no
proper subset can be sufficient, hence the output after the minimality
filter is precisely \(P_{\min}(x,y)\).}

\section{Computing Global Explanations}
\hspace{-1mm}
\begin{figure}
    \centering
    \includegraphics[width=0.3\linewidth]{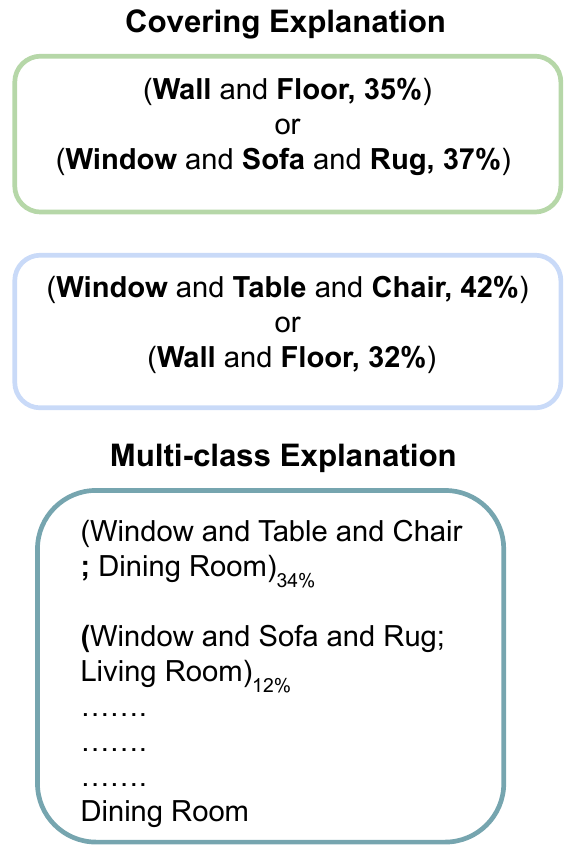}
        \includegraphics[width=0.6\linewidth]{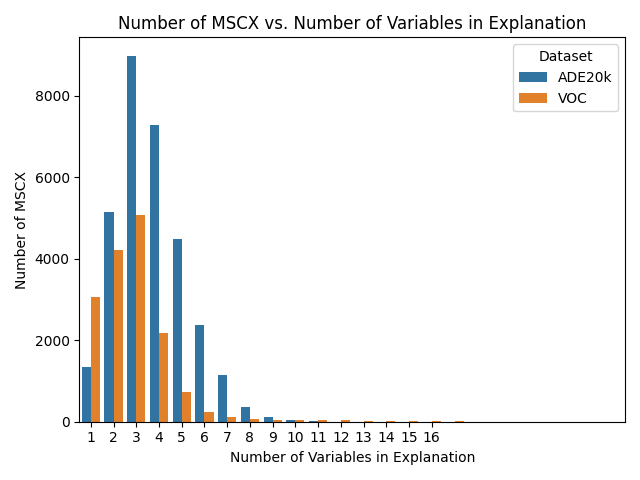}
    \caption{ Local explanations across multiple images are used to generate global covering explanations that cover a given class i.e., Covering explanations (left-top) vs Multi-class Explanations (left-bottom) that explain all classes. (Right) The size (number of literals) of the MSCX vs the number of MSCXs of that size.}
    \label{fig:num_vars}
\end{figure}

For every image \(x\) with predicted class \(y\) we collect the set of all its MSCXs represented as 
$P_{\min}(x,y)$. 
A multi-class global explanation can be viewed as a 
sequence of clauses.
Two concrete realizations used in this paper are:

\begin{enumerate}[label=(\roman*),nosep,leftmargin=*]
\item \textbf{Covering (Alg.\,\ref{alg:greedyMDNF}).}
      The sequence is \emph{unordered}; the disjunction
      \(\Phi_{y}(x)=\bigvee_{j}\bigwedge_{o\in S_{j}}z_{o}(x)\)
      must satisfy all instances $\mathcal I_{y}$ which are predicted to be in class $y$. That is,   
      \(
        \forall x\in\mathcal I_{y}:\;\Phi_{y}(x)=1.
      \)

\item \textbf{Multi-Class (Alg.\,\ref{alg:decision-list}).}
      The sequence is an \emph{ordered} explanation list, where each clause in the list is a pair of MSCX and a class. The first MSCX that explains an instance

      \[
    \text{(}S_{1};c_{1}\text{)}_{d_1\%}
    \;\prec\;
    \dots
    \;\prec\;
    \text{(}S_{N-1};c_{N-1}\text{)}_{d_{N-1\%}}
    \;\prec\;
    \text{(}\varnothing;c_N\text{)}_{d_N}
\]
The explanation list can be read as: The dataset is partitioned into $N$ groups, each of size $d_i$\%, where partition $i$ is explained by MSCX $S_i$ to be in class $c_i$ and not explained by any previous $S_j$ where $j <i$.

\end{enumerate}

The ideal global explanation for a class $y$ must satisfy the following conditions. 
\begin{enumerate}[label=(C\arabic*), leftmargin=*]
\item \textbf{Local sufficiency.}  
      Every antecedent mask \(S_{j}\) originates from some
      \(P_{\min}(x,y)\); therefore
      \(f_{y}(x_{S_{j}})\ge\tau_{P}f_{y}(x)\) on the images it covers.

\item \textbf{Exact coverage.}  
      Each image in \(\mathcal I_{y}\) must be covered by
      at least one clause in the covering explanation or must be classified correctly by the 
      explanation list.

\item \textbf{Parsimony.}  All else being equal the sizes of the covering explanation and the explanation list must be as small as possible. 
     
\end{enumerate}

Unfortunately, selecting the smallest subset of MSCXs
\(\mathcal T\subseteq\mathcal P_{y}\)
satisfying exact coverage (C2) 
is 
NP-hard and inapproximable within 
\((1-\tfrac1e)\ln|\mathcal I_{y}|\) unless
\(\mathbf P=\mathbf{NP}\) \cite{feige1998threshold}.
An exhaustive solver must evaluate  all 
\(2^{|\mathcal P_{y}|}-1\) subsets.  
Minimising the rule list’s empirical error subject to one-pass coverage
is equivalent to the NP-complete \emph{Minimum Consistent Decision List}
problem \cite{rivest1987learning}.

To avoid the exponential complexity of exhaustive approaches, we use greedy strategies to arrive at approximate solutions that are still effective. Both our strategies honor Conditions (C1)–(C2) by construction and run in polynomial time, but due to the approximation from using Greedy approaches we cannot guarantee (C3).

\subsection{Finding Covering Explanations}

For a covering global explanation, we use the greedy set cover algorithm to compute a monotone DNF formula, where each term in the DNF is an MSCX (C1) and each example is explained (covered) by at least one MSCX (C2).

We first construct $P_{min}(x, y)$ for all images $x$ in the dataset that are classified as $y$. At each iteration of our algorithm, we  find an MSCX $S*$ that maximizes the number of instances it explains from $U$. The term found is added to the DNF as a clause and all instances covered by the clause are removed from the pool $U$.  This process is repeated till no more images have to be covered. For formally, fix class \(y\) and denote by
\(\mathcal I_{y}=\{x^{(n)}\mid f(x^{(n)})=y\}\) the  images of
that class.  For every image we already computed its 
MSCXs
\(P_{\min}(x,y)\).  The greedy routine below iteratively chooses the
$S*$ that explains the \emph{largest} number of still-uncovered images
and appends it as a new clause in the explanation.
\begin{algorithm}
\caption{\textsc{GreedyCover}$(\mathcal I_{y},
                               \{P_{\min}(x,y)\})$}
\label{alg:greedyMDNF}
\begin{algorithmic}[1]
\Require support images \(\mathcal I_{y}\) and their
         \(P_{\min}(x,y)\)
\Ensure  ordered clause list \(\mathcal T\)
\vspace{1pt}
\State build map
      \(
        \mathcal R:S\mapsto
        \{x\in\mathcal I_{y}\,|\,S\in P_{\min}(x,y)\}
      \)
      \Comment{mask \(\to\) images it covers}
\State $U\gets\mathcal I_{y}$,\;
      \(\mathcal T\gets\varnothing\) \Comment{uncovered set and clause list}
\While{$U\neq\varnothing$}
    \State select
          \(S^{\star}=
               \displaystyle\arg\max_{S\in\mathcal R}
               |\mathcal R(S)\cap U|\)
          \Comment{max‐gain MSCX}
    \If{$\mathcal R(S^{\star})\cap U=\varnothing$}
        \State \textbf{break} \Comment{no mask covers a new image}
    \EndIf
    \State $\mathcal T\gets\mathcal T\cup\{S^{\star}\}$ \Comment{add clause}
    \State $U\gets U\setminus\mathcal R(S^{\star})$ \Comment{remove covered images}
\EndWhile
\State \Return $\mathcal T$
\end{algorithmic}
\end{algorithm}

Given the clause set $\mathcal T$ returned by
Algorithm~\ref{alg:greedyMDNF}, the global explanation for class~$y$ is represented  as an unordered monotone DNF as shown below.
\begin{align*}
   \Phi_{y}(x)
     \;=\;
     \bigvee_{S\in\mathcal T}
       \Bigl(\,\bigwedge_{o\in S} z_{o}(x)\Bigr)_{d_S\%},
\end{align*}
where $z_{o}(x)$ indicates the presence of object $o$ in image $x$ and $d_S = \frac{|\mathcal R(S)|}{\mathcal I_y}$. Let \(M=|\mathcal R|\) be the number of distinct minimal masks and
\(N=|\mathcal I_{y}|\).
Computing \(|\mathcal R(S)\cap U|\) for every \(S\) each round is
\(O(M)\); with at most \(N\) iterations the overall cost is
\(O(MN)\).  
Upon termination every $x \in \mathcal I_y$ is covered by at least one clause of ~\(\Phi_{y}\).

\begin{theorem}
If $P_{min}(x,y) \not = \varnothing$ is a complete explanation of each image $x$ in the set $\mathcal I_y$, 
and $m*$ is the minimum number of MSCXs needed to explain (cover) all images in $\mathcal I_y$, then Algorithm 2 outputs a global explanation of size at most $(m^{*} \ln |\mathcal I_y|)$.
\end{theorem}

{\em Proof:} Since each instance has at least one term that explains it in $P_{min}(x,y)$, every example in $\mathcal I_y$ will be covered by the final output. From the approximation bound of the greedy set cover, it follows that $(m^{*} \log |\mathcal I_y|)$ terms are sufficient to fully cover the images in $\mathcal I_y$  \cite{feige1998threshold}. $\Box$

\subsection{Global Multi-Class Explanations}

One problem with covering explanations is that they explain each class separately. Unfortunately this could result in multiple overlapping explanations for different classes, which is not ideal and may even be confusing. 
Multi-class explanations seek to explain multiple classes 
simultaneously in the form of an `explanation list' with interleaved but non-overlapping explanations.  

\begin{algorithm}[H]
\caption{\textsc{ExplanationList}\((\mathcal X, \mathcal Y \in \mathcal{D},
                                \{P_{\min}(x,y)\}_{x\in \mathcal X, y\in \mathcal Y})\)}
\label{alg:decision-list}
\begin{algorithmic}[1]
\Require
  set of image-label pair \( \mathcal X, \mathcal Y \in \mathcal D\) and 
\Require
  their minimal masks   \(P_{\min}(x,y)\)
\Ensure
  ordered list of rules \(\mathcal L\) in the form 
  “IF pattern THEN class”, plus a final ELSE

\vspace{2pt}
\State $U\gets\mathcal X$ \Comment{images not yet covered}
\State $\mathcal C\gets\varnothing$ \Comment{explanation-list \(\mathcal L\)}
\State build dictionary
      \( \mathcal M = \bigl\{\,S\mapsto
           \{\,(x,c)\mid x\in\mathcal D,
                S\in P_{\min}(x,y),\,
                c=\operatorname{pred}(x)\}\bigr\} \)

\While{$U\neq\varnothing$ \textbf{and} \(\exists S\in\mathcal M\)
       s.t.\ \(S\) covers a new image}
    \ForAll{$S\in\mathcal M$}
        \State compute
          \(N_{S,c} \;:=\;
            \bigl\{x\in U\mid (x,c)\in\mathcal M[S]\bigr\}\)
          for every class \(c\)
        \State $T_{S}\gets\bigcup_{c}N_{S,c}$ \Comment{new images by \(S\)}
        \If{$T_{S}=\varnothing$}\ \textbf{continue}
        \State $c^{\star}_{S}\gets
               \arg\max_{c}|N_{S,c}|$  \Comment{majority class}
        \State $\mathrm{err}_{S}\gets|T_{S}|-|N_{S,c^{\star}_{S}}|$
        \State $\mathrm{gain}_{S}\gets|T_{S}|$
        \EndIf
    \EndFor
    \State choose 
      \(S^{\star}=\displaystyle
        \arg\min_{S}\bigl(\mathrm{err}_{S},-\mathrm{gain}_{S}\bigr)\)
    \State append rule \((S^{\star};c^{\star}_{S^{\star}})\)
      to \(\mathcal C\)
    \State $U\gets U\setminus T_{S^{\star}}$ 
           \Comment{mark images as covered}
    \State remove \(S^{\star}\) from \(\mathcal M\)
\EndWhile
\State append default rule 
      \(\varnothing\operatorname*{mode}_{x\in\mathcal D}
        \operatorname{pred}(x)\)
      to \(\mathcal C\)
\State \Return \(\mathcal C\)
\end{algorithmic}
\end{algorithm}

\subsubsection*{Description of Algorithm \ref{alg:decision-list}}

\begin{enumerate}[label=\textbf{Step \arabic*:}, leftmargin=*]
\item \emph{Initialisation}  
      The set \(U\subseteq\mathcal D\) holds all images that still
      need an explanation; the explanation list \(\mathcal C\) is empty.

\item \emph{Index minimal masks.}  
      For every distinct MSCX \(S\) that appears in any
      \(P_{\min}(x,y)\) we store the bag  
      \(\mathcal M[S]=\{(x,c)\}\), containing each image \(x\) in which
      \(S\) occurs together with its predicted class
      \(c=\operatorname{pred}(x)\).
      This hash map lets us query, in \(O(1)\), which images a mask
      covers and what labels they receive.

\item \emph{Iterative rule extraction.}
      While uncovered images remain and at least one MSCX still covers a
      new image we

      \begin{enumerate}[label*=\arabic*.]
      \item count, for every mask \(S\), how many \emph{new} images it
            covers per class (\(N_{S,c}\));  
      \item compute its total new coverage
            \(|T_{S}|=\sum_{c}|N_{S,c}|\) and the associated error
            \(\mathrm{err}_{S}=|T_{S}|-\max_{c}|N_{S,c}|\);  
      \item pick the mask \(S^{\star}\) with the smallest error, breaking
            ties by the largest gain \(\mathrm{gain}_{S}=|T_{S}|\);  
      \item append the rule
            \((S^{\star};
              c^{\star})=\arg\max_{c}|N_{S^{\star},c}|\)
            to \(\mathcal C\);  
      \item mark all images in \(T_{S^{\star}}\) as covered
            (\(U\gets U\setminus T_{S^{\star}}\)) and delete
            \(S^{\star}\) from the mask dictionary so it cannot be reused.
      \end{enumerate}

      This greedy choice minimises newly induced errors at each round
      while maximising the number of newly explained images.

\item \emph{Default rule.}  
      Once no mask can cover a fresh image or \(U=\varnothing\),  
      we add a terminal clause whose conclusion is the majority
      class in \(\mathcal D\).
      Any image that slipped through the mask rules is now assigned a
      class, guaranteeing completeness.

\item \emph{Return the explanation list.}  
      The ordered rule set \(\mathcal C\) forms a monotone explanation list, where 
      the first antecedent that matches an input determines its class.
 
\end{enumerate}

\begin{theorem}
If there exists a multi-class explanation list over 
the terms (MSCXs) in ${\cal T} = \{\bigcup P_{min}(x,y) | x\in \mathcal D, y = argmax_y f_y(x) \}$ with 100\% accuracy on $\mathcal D$, then Algorithm 3 will output a (possibly different) explanation list of at most size $\min (|{\cal T}|, |\mathcal D|)$ and 100\% accuracy. 
\end{theorem}

\remove{
{\em Proof:} The proof combines the ideas behind 
the decision list algorithm of \cite{rivest1987learning}
and the proof of approximation bound of greedy set cover \cite{feige1998threshold}. 
Assume that just before the $j^{th}$ iteration of the while loop of Algorithm 3, there are $x_j$ images yet to be covered. Since $O$ covers all these images $|O| = m*$, there must exist one clause which covers at least $\frac {x_j} {m*}$ images. 
Further assume that the selected clause in 
that iteration covers $n_j$ of these images. 
$n_j \geq \frac {x_j} {m*}$ if the selection is purely greedy according to coverage. 
{\bf Problem:  But we also need one with zero errors. }}

{\em Proof:} Assume that the target explanation list consists of 
clauses $C_1, \ldots C_n$, and that Algorithm 3
has selected clauses $C_1, \ldots, C_{i-1}$ without errors and selected a different clause $C_i' \not = C_i$. 
Since selecting $C_i$ would not have given any errors, 
it follows that $C_i'$ has at least as much additional coverage as $C_i$ and no errors. 
The remaining examples can still be covered by the 
explanation list $C_i, \ldots, C_{n}$ without errors, since they are all in the original example set. 
Hence it will never face a situation where there is no clause to cover at least one new example unless all clauses in the target are used. By assumption, then we would have 100\% coverage. Since each clause occurs at most once in the explanation list and covers at least one instance in $D$, the number of clauses is bounded by $Min(|{\cal T}|,|D|)$. $\Box$ \\
\emph{Complexity:} Let $M$ be the size of ${\cal T}$, the 
union of all MSCXs for all images and their predicted classes. 
Each iteration scans every still-available mask and touches
each mask–image pair at most once, while at least one new image is
removed from the uncovered pool.  Consequently the entire procedure
runs in 
$O\!\bigl(M^{2}+M|D|\bigr)$ time and $ O( M|D|)$ space.  In practice 
\(M\ll |D|\), so the algorithm is close to linear in the dataset size
while avoiding the factorial blow-up of an optimal explanation-list search.

\section{Experimental Setup and Results}

We assess the proposed method on two fronts: \textit{local} quality, via $Fiedility^{+}$ and $Fiedility^{-}$, and \textit{global} quality of the MDNF via the coverage of the induced rules over the validation set and quality of multi-class explanations via validation accuracy.  Experiments are conducted on ADE20K\cite{zhou2019semantic} and Pascal-Parts\cite{chen2014detect}. ADE20K exhibits a pronounced long-tail; classes containing fewer than 15 images are excluded, leaving the ten most populous categories with annotations for objects in the scene.  Pascal-Parts is used in its entirety along with annotations for parts of an object demonstrating the adaptability of our approach at a human-specified granularity. To probe architectural robustness, we compare a feed-forward convolutional network (VGG-19 \cite{SimonyanZisserman2015}) with a Vision Transformer backbone (ViT-B \cite{dosovitskiyimage}). Both models were trained for 100 epochs with a fixed learning rate of 0.001; we did not perform any hyperparameter tuning, as optimizing model performance was not our goal.  On average, each model attained 90 \% validation accuracy.  In Algorithm \ref{alg:beam-add}, we used a beam width $B = 3$ and generated up to 5 successors per expansion.
\begin{table}
\centering
\caption{Fidelity scores (\textit{mean$\pm$std}) for covering MDNFs on ADE20K and Pascal-Parts, evaluated with VGG-19 and ViT backbones. Higher Fid$^{-}$ and lower Fid$^{+}$ are better. Validation Accuracy indicates the ability of Multi-class explanation list in replicating model behavior.}
\small
\begin{tabular}[t]{llccc}
\toprule
\multirow{2}{*}{\textbf{Dataset}} & \multirow{2}{*}{\textbf{Architecture}}
  & \multicolumn{1}{c}{\textbf{Multi-class}} & \multicolumn{2}{c}{\textbf{Covering}} \\
\cmidrule(lr){3-3}\cmidrule(l){4-5}
 & & Accuracy & Fid$^{+} (\downarrow)$ & Fid$^{-} (\uparrow)$ \\
\midrule
\multirow{2}{*}{ADE20K}
  & VGG-19 & 73.51 & 0.113$\pm$0.297 & 0.992$\pm$0.090 \\
  & ViT-B  & 75.69 & 0.314$\pm$0.378 & 1.029$\pm$0.173 \\
\midrule
\multirow{2}{*}{Pascal-Parts}
  & VGG-19 & 50.32 & 0.219$\pm$0.201 & 0.961$\pm$ 0.054 \\
  & ViT-B  & 53.49 & 0.339$\pm$0.35 & 0.955$\pm$0.041 \\
\bottomrule
\end{tabular}
\end{table}
\par  We present two versions of our approach: 

(1) \textbf{Covering} explanations that focus solely on covering the target class. 
and, (2)  \textbf{Multi-class} explanations that compile explanations for all classes into an explanation list that prioritizes MSCXs that maximize coverage. Since the Covering explanation aims at explaining the classifications into a single class, we use coverage as the evaluation metric. Coverage measure how many of the validation decisions are explained by the different MSXs in the global explanation of images from a single class. For the multi-class case, we report the validation accuracy of the explanation list to match the classifier's decision. A high accuracy in this context
(which should not be confused with classification accuracy), 
demonstrates that our explanation list faithfully explains the 
model's decision making in terms of the relevant objects.

\begin{figure}
    \centering
    \includegraphics[width=0.49\linewidth]{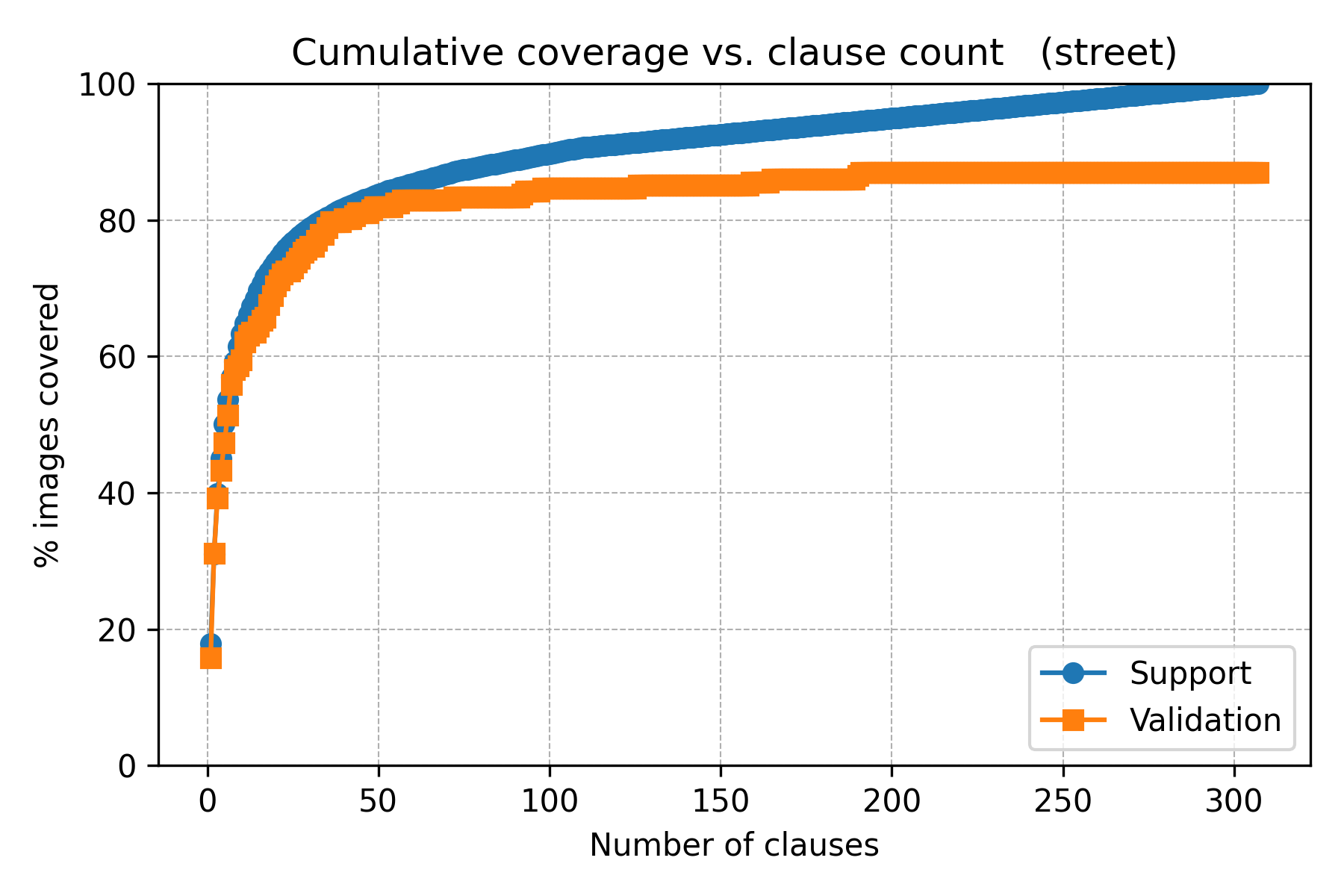}
    \includegraphics[width=0.49\linewidth]{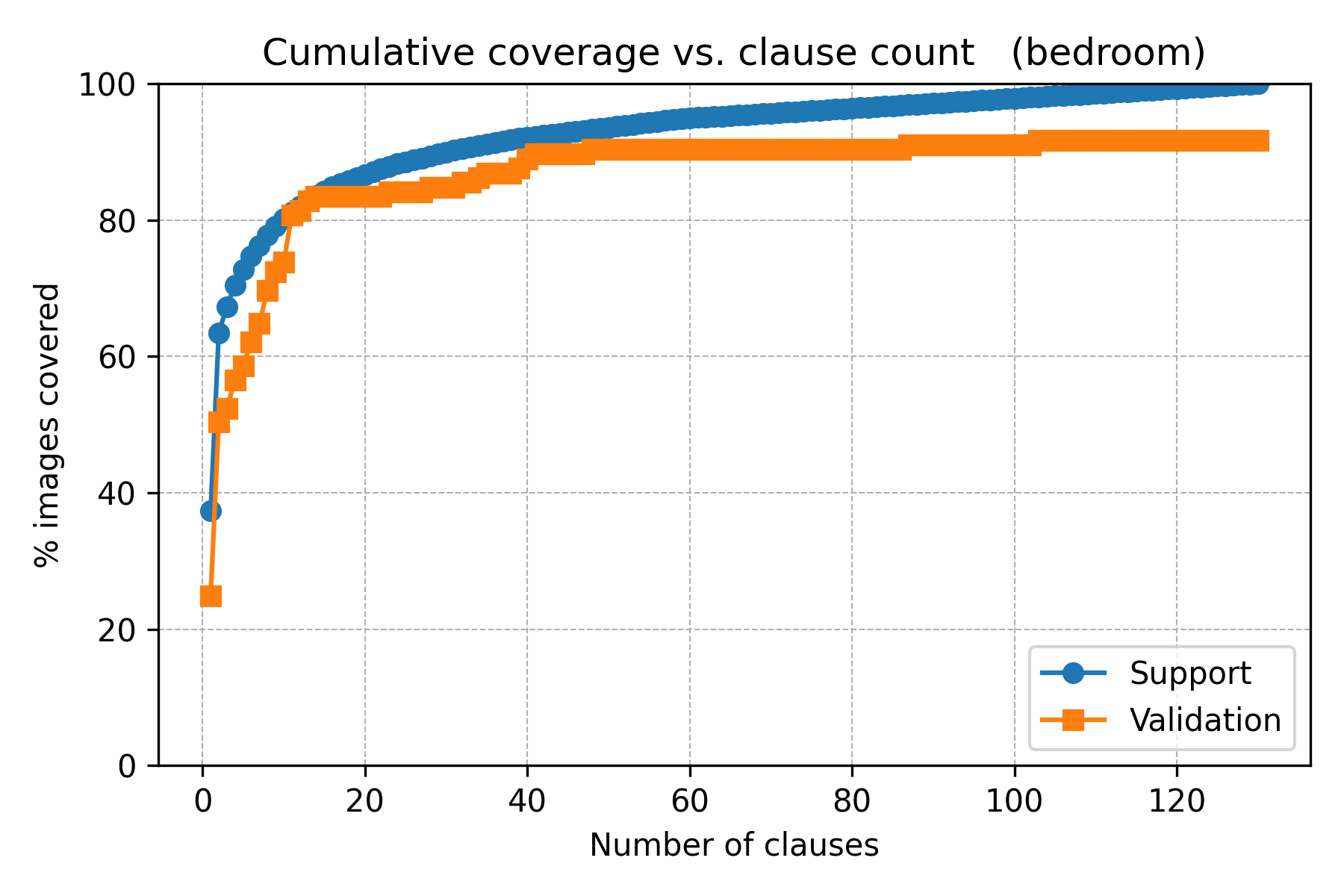}
 
    \caption{Percentage of images covered vs. 
    number of clauses in the covering explanations for class 'Street' and 'Bedroom' from ADE20k dataset.}
    \label{fig:coverage_plots}
\end{figure}

Figure \ref{fig:num_vars} plots the distribution of the number of concepts per MSCX, which closely approximates a Gaussian distribution centered at three concepts per MSCX. Table 1 reports the mean $\pm$ std of Fidelity$^{+}$ and Fidelity$^{-}$ on the validation set, together with the Covering MDNF’s multi‐class explanation‐list accuracy. Fidelity$^{+}$ is the ratio of the model’s score after removing all MSCX concepts to the original score; Fidelity$^{-}$ is the ratio when retaining only one MSCX. Because removing a strong cross‐class indicator can increase the focal class’s score, Fidelity$^{-}$ may exceed 1. The ViT‐B backbone achieves substantially higher Fidelity$^{+}$, indicating its richer representations yield decisions only partly captured by the MDNF; extracting additional MSCX concepts could recover the remainder.
Finally, the multi‐class explanation list achieves a validation accuracy of 75.69\% for the ViT‐B model on ADE20K while the accuracy drops to 53.4\% on the Pascal-Parts dataset. Given the model’s native accuracy of approximately 90\%, this corresponds to explaining about 85\% of its behavior.

For illustration, we show below  
the global explanation for the class ``Bedroom” with atleast 3\% coverage.
\begin{align*}
\Phi_{\text{Bedroom}} = &(\text{bed})_{\;26\%} \lor(\text{wall} \land \text{bed})_{\;25\%}\lor (\text{wall} \land \text{Background})_{\;7\%} \\ &\lor(\text{wall} \land \text{sofa})_{\;5\%} \lor(\text{wall})_{\;4\%} \lor(\text{wall} \land \text{floor} \land \text{bed})_{\;3\%} 
\end{align*}

Unlike the local explanations, global explanations have a different interpretation because they are constructed out of MSCXs from different examples. For example,  
the above explanation suggests that 26\% of the images of bedrooms are explained by the presence of \text{bed}
and another 25\% of the images of bedrooms are explained by the presence of \text{bed} and \text{wall}. Unlike in  logical formulae, \text{bed} here does not subsume 
\text{bed} and \text{wall} because in 25\% examples 
\text{bed} alone is not minimally sufficient. Figure \ref{fig:coverage_plots} shows support and validation‐set coverage versus the number of MDNF clauses for the classes “Street” and “Bedroom”. 
By design, support coverage reaches 100 \%, and its steep early rise highlights clauses that explain large class fractions.  On the validation set, coverage plateaus at approximately 85\% with only 20 clauses, demonstrating that the greedy selection effectively identifies the most influential terms.

\section{Conclusion and Future Work}
We have presented two complementary strategies for turning local,
$\tau_{P}$-minimal image based explanations into global, human-readable explanations. Our \textbf{covering-MDNF} converts the complete explanations 
for a class into a compact monotone DNF explanation 
which covers about 85\% of the validation decisions with as little as 20 clauses.
  For multi-class settings we proposed a \textbf{greedy explanation list} that orders the same MSCX to minimize newly introduced errors while retaining the instance-level sufficiency guarantee. Empirically they produce explanations an order of magnitude
smaller than the dataset they summarize, making deep classifiers
substantially more transparent without retraining. Future work will spearhead exploration in the following directions: \textbf{(1) Object removal via tractable generative models:} Instead of Gaussian blurring, we will employ tractable probabilistic models to inpaint or erase objects \cite{liuimage}, mitigating the out-of-distribution artifacts that might plague occlusion-based perturbations.\textbf{(2) Temporal generalization:} Extending the framework to video will test both the temporal stability of local explanations and the scalability of global rule synthesis across frames.

\bibliographystyle{splncs04}

\bibliography{references}

@incollection{karp2009reducibility,
  title={Reducibility among combinatorial problems},
  author={Karp, Richard M},
  booktitle={50 Years of Integer Programming 1958-2008: from the Early Years to the State-of-the-Art},
  pages={219--241},
  year={2009},
  publisher={Springer}
}

@article{feige1998threshold,
  title={A threshold of ln n for approximating set cover},
  author={Feige, Uriel},
  journal={Journal of the ACM (JACM)},
  volume={45},
  number={4},
  pages={634--652},
  year={1998},
  publisher={ACM New York, NY, USA}
}

@article{carion2025sam,
  title={SAM 3: Segment Anything with Concepts},
  author={Carion, Nicolas and Gustafson, Laura and Hu, Yuan-Ting and Debnath, Shoubhik and Hu, Ronghang and Suris, Didac and Ryali, Chaitanya and Alwala, Kalyan Vasudev and Khedr, Haitham and Huang, Andrew and others},
  journal={arXiv preprint arXiv:2511.16719},
  year={2025}
}

@article{rivest1987learning,
  title={Learning decision lists},
  author={Rivest, Ronald L},
  journal={Machine learning},
  volume={2},
  pages={229--246},
  year={1987},
  publisher={Springer}
}

@article{vasu2025beyond,
  title={Beyond Local Explanations: A Framework for Global Concept-Based Interpretation in Image Classification},
  author={Vasu, Bhavan and Rathore, Kunal and Tadepalli, Prasad},
  journal={Electronics},
  volume={14},
  number={16},
  pages={3230},
  year={2025},
  publisher={MDPI}
}

@article{selvaraju2020grad,
  title={Grad-CAM: visual explanations from deep networks via gradient-based localization},
  author={Selvaraju, Ramprasaath R and Cogswell, Michael and Das, Abhishek and Vedantam, Ramakrishna and Parikh, Devi and Batra, Dhruv},
  journal={International journal of computer vision},
  volume={128},
  pages={336--359},
  year={2020},
  publisher={Springer}
}

@article{baugh2025neural,
  title={Neural DNF-MT: A Neuro-symbolic Approach for Learning Interpretable and Editable Policies},
  author={Baugh, Kexin Gu and Dickens, Luke and Russo, Alessandra},
  journal={arXiv preprint arXiv:2501.03888},
  year={2025}
}

@inproceedings{koh2020concept,
  title={Concept bottleneck models},
  author={Koh, Pang Wei and Nguyen, Thao and Tang, Yew Siang and Mussmann, Stephen and Pierson, Emma and Kim, Been and Liang, Percy},
  booktitle={International conference on machine learning},
  pages={5338--5348},
  year={2020},
  organization={PMLR}
}

@inproceedings{xulogicmp,
  title={LogicMP: A Neuro-symbolic Approach for Encoding First-order Logic Constraints},
  author={Xu, Weidi and Wang, Jingwei and Xie, Lele and He, Jianshan and Zhou, Hongting and Wang, Taifeng and Wan, Xiaopei and Chen, Jingdong and Qu, Chao and Chu, Wei},
  booktitle={The Twelfth International Conference on Learning Representations},
  year={2024},
}

@inproceedings{fischer2019dl2,
  title={DL2: training and querying neural networks with logic},
  author={Fischer, Marc and Balunovic, Mislav and Drachsler-Cohen, Dana and Gehr, Timon and Zhang, Ce and Vechev, Martin},
  booktitle={International Conference on Machine Learning},
  pages={1931--1941},
  year={2019},
  organization={PMLR}
}

@inproceedings{darwiche2022computation,
  title={On the computation of necessary and sufficient explanations},
  author={Darwiche, Adnan and Ji, Chunxi},
  booktitle={Proceedings of the AAAI Conference on Artificial Intelligence},
  volume={36},
  pages={5582--5591},
  year={2022}
}

@inproceedings{dosovitskiyimage,
  title={An Image is Worth 16x16 Words: Transformers for Image Recognition at Scale},
  author={Dosovitskiy, Alexey and Beyer, Lucas and Kolesnikov, Alexander and Weissenborn, Dirk and Zhai, Xiaohua and Unterthiner, Thomas and Dehghani, Mostafa and Minderer, Matthias and Heigold, Georg and Gelly, Sylvain and others},
  booktitle={International Conference on Learning Representations},
  year={2021}

}

@inproceedings{SimonyanZisserman2015,
  author       = {Simonyan, Karen and Zisserman, Andrew},
  title        = {Very Deep Convolutional Networks for Large-Scale Image Recognition},
  booktitle    = {3rd International Conference on Learning Representations (ICLR 2015)},
  organization = {Computational and Biological Learning Society},
  year         = {2015},
  pages        = {1--14},
}

@inproceedings{chen2014detect,
  title={Detect what you can: Detecting and representing objects using holistic models and body parts},
  author={Chen, Xianjie and Mottaghi, Roozbeh and Liu, Xiaobai and Fidler, Sanja and Urtasun, Raquel and Yuille, Alan},
  booktitle={Proceedings of the IEEE conference on computer vision and pattern recognition},
  pages={1971--1978},
  year={2014}
}

@article{zhou2019semantic,
  title={Semantic understanding of scenes through the ade20k dataset},
  author={Zhou, Bolei and Zhao, Hang and Puig, Xavier and Xiao, Tete and Fidler, Sanja and Barriuso, Adela and Torralba, Antonio},
  journal={International Journal of Computer Vision},
  volume={127},
  number={3},
  pages={302--321},
  year={2019},
  publisher={Springer}
}

@inproceedings{bau2017network,
  title={Network dissection: Quantifying interpretability of deep visual representations},
  author={Bau, David and Zhou, Bolei and Khosla, Aditya and Oliva, Aude and Torralba, Antonio},
  booktitle={Proceedings of the IEEE conference on computer vision and pattern recognition},
  pages={6541--6549},
  year={2017}
}

@inproceedings{fong2017interpretable,
  title={Interpretable explanations of black boxes by meaningful perturbation},
  author={Fong, Ruth C and Vedaldi, Andrea},
  booktitle={Proceedings of the IEEE international conference on computer vision},
  pages={3429--3437},
  year={2017}
}

@inproceedings{liuimage,
  title={Image Inpainting via Tractable Steering of Diffusion Models},
  author={Liu, Anji and Niepert, Mathias and Van den Broeck, Guy},
  booktitle={The Twelfth International Conference on Learning Representations},
  year={2024}

}

@inproceedings{azzolin2023global,
  title={Global Explainability of GNNs via Logic Combination of Learned Concepts},
  author={Azzolin, Steve and Longa, Antonio and Barbiero, Pietro and Lio, Pietro and Passerini, Andrea and others},
  booktitle={11th International Conference on Learning Representations (ICLR 2023)},
  pages={1--19},
  year={2023},
  organization={International Conference on Learning Representations (ICLR)}
}

@article{craven1995extracting,
  title={Extracting tree-structured representations of trained networks},
  author={Craven, Mark and Shavlik, Jude},
  journal={Advances in neural information processing systems},
  volume={8},
  year={1995}
}

@inproceedings{zhou2016learning,
  title={Learning deep features for discriminative localization},
  author={Zhou, Bolei and Khosla, Aditya and Lapedriza, Agata and Oliva, Aude and Torralba, Antonio},
  booktitle={Proceedings of the IEEE conference on computer vision and pattern recognition},
  pages={2921--2929},
  year={2016}
}

@inproceedings{DBLP:conf/bmvc/PetsiukDS18,
  author       = {Vitali Petsiuk and
                  Abir Das and
                  Kate Saenko},
  title        = {{RISE:} Randomized Input Sampling for Explanation of Black-box Models},
  booktitle    = {British Machine Vision Conference 2018, {BMVC} 2018, Newcastle, UK,
                  September 3-6, 2018},
  pages        = {151},
  publisher    = {{BMVA} Press},
  year         = {2018},
  url          = {http://bmvc2018.org/contents/papers/1064.pdf},
  bibsource    = {dblp computer science bibliography, https://dblp.org}
}

@article{zhang2019should,
  title={" Why should you trust my explanation?" understanding uncertainty in LIME explanations},
  author={Zhang, Yujia and Song, Kuangyan and Sun, Yiming and Tan, Sarah and Udell, Madeleine},
  journal={arXiv preprint arXiv:1904.12991},
  year={2019}
}

@article{shitole2021one,
  title={One explanation is not enough: structured attention graphs for image classification},
  author={Shitole, Vivswan and Li, Fuxin and Kahng, Minsuk and Tadepalli, Prasad and Fern, Alan},
  journal={Advances in Neural Information Processing Systems},
  volume={34},
  pages={11352--11363},
  year={2021}
}

@inproceedings{mosca2022shap,
  title={SHAP-based explanation methods: a review for NLP interpretability},
  author={Mosca, Edoardo and Szigeti, Ferenc and Tragianni, Stella and Gallagher, Daniel and Groh, Georg},
  booktitle={Proceedings of the 29th international conference on computational linguistics},
  pages={4593--4603},
  year={2022}
}

@inproceedings{abasi2014exact,
  title={On exact learning monotone DNF from membership queries},
  author={Abasi, Hasan and Bshouty, Nader H and Mazzawi, Hanna},
  booktitle={International Conference on Algorithmic Learning Theory},
  pages={111--124},
  year={2014},
  organization={Springer}
}

\end{document}